\documentclass[11pt,a4paper]{article}
\usepackage[hyperref]{acl2018}
\usepackage{times}
\usepackage{latexsym}
\usepackage{graphicx}
\usepackage{subfigure}
\usepackage{amsmath,amssymb,amsfonts,comment}
\usepackage{multirow}
\usepackage{colortbl}

\usepackage{url}

\newcommand{\specialcell}[2][c]{%
  \begin{tabular}[#1]{@{}l@{}}#2\end{tabular}}

\DeclareMathOperator*{\gru}{GRU}
\DeclareMathOperator*{\softmax}{Softmax}

\aclfinalcopy 

\title{Working Memory Networks: Augmenting Memory Networks with a Relational Reasoning Module}

\author{Juan Pavez*,  H\'ector Allende \\
  Department of Informatics \\ Federico Santa Mar\'ia \\ Technical University \\ Valpara\'iso, Chile \\
  {\tt juan.pavezs@alumnos.usm.cl}\\
  {\tt hallende@inf.utfsm.cl} \\\And
  H\'ector Allende-Cid \\
  Escuela de Ingenier\'ia Inform\'atica \\ Pont\'ifica Universidad Cat\'olica \\ de Valpara\'iso \\  Valpara\'iso, Chile \\
  {\tt hector.allende@pucv.cl} \\}

\date{}

\begin{document}
\maketitle
\begin{abstract}
During the last years, there has been a lot of interest in achieving some kind of complex reasoning using deep neural networks. To do that, models like Memory Networks (MemNNs) have combined external memory storages and attention mechanisms. These architectures, however, lack of more complex reasoning mechanisms that could allow, for instance, relational reasoning. Relation Networks (RNs), on the other hand, have shown outstanding results in relational reasoning tasks. Unfortunately, their computational cost grows quadratically with the number of memories, something prohibitive for larger problems. To solve these issues, we introduce the Working Memory Network, a MemNN architecture with a novel working memory storage and reasoning module. Our model retains the relational reasoning abilities of the RN while reducing its computational complexity from quadratic to linear. We tested our model on the text QA dataset bAbI and the visual QA dataset NLVR. In the jointly trained bAbI-10k, we set a new state-of-the-art, achieving a mean error of less than 0.5\%. Moreover, a simple ensemble of two of our models solves all 20 tasks in the joint version of the benchmark.

\end{abstract}

\section{Introduction}

A central ability needed to solve daily tasks is complex reasoning. It involves the capacity to comprehend and represent the environment, retain information from past experiences, and solve problems based on the stored information. Our ability to solve those problems is supported by multiple specialized components, including short-term memory storage, long-term semantic and procedural memory, and an executive controller that, among others, controls the attention over memories \cite{baddeley1992working}.

\begin{figure*}[ht]
\vskip 0.2in
\begin{center}
\centerline{\includegraphics[width=1.8\columnwidth]{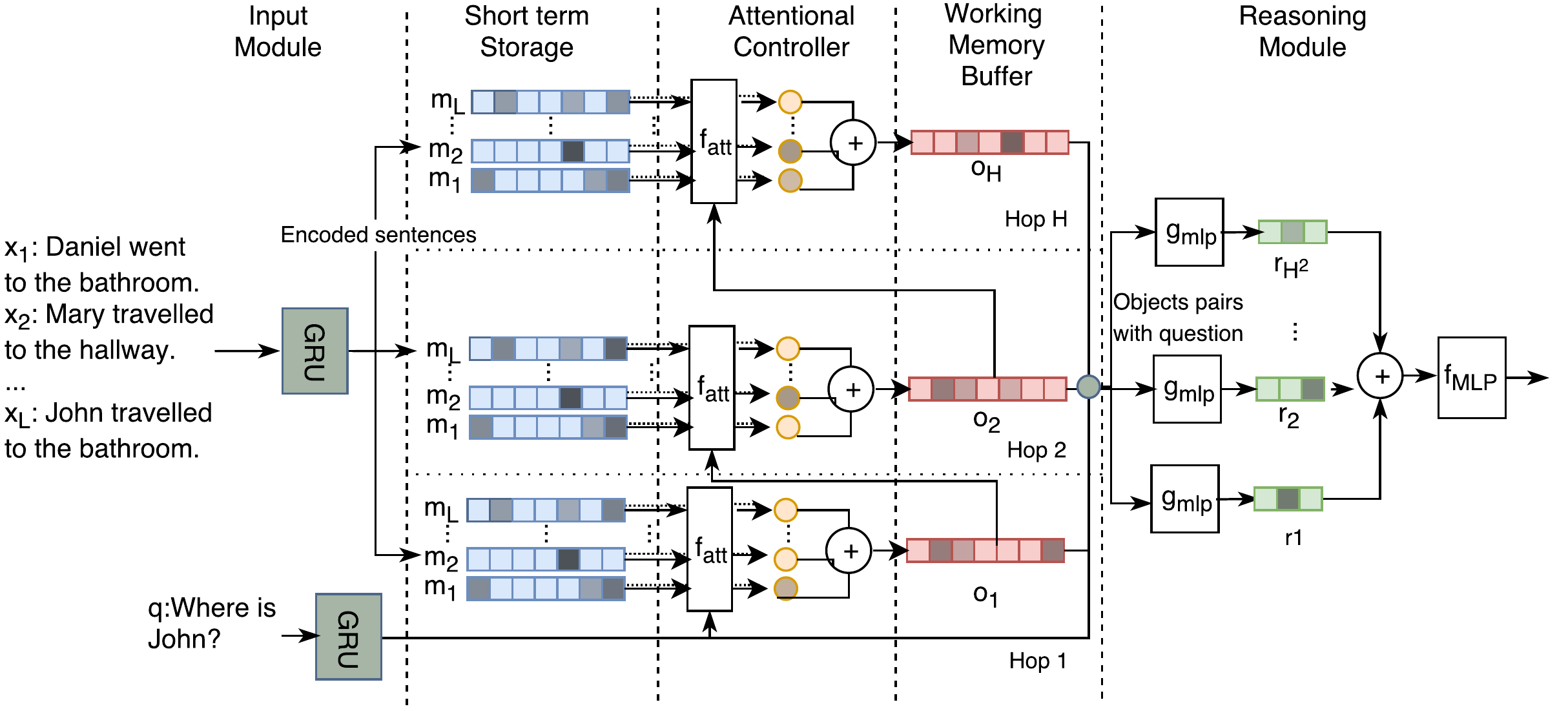}}
\caption{The W-MemNN model applied to textual question answering. Each input fact is processed using a GRU, and the output representation is stored in the short-term memory storage. Then, the attentional controller computes an output vector that summarizes relevant parts of the memories. This process is repeated $H$ hops (a dotted line delimits each hop), and each output is stored in the working memory buffer. Finally, the output of each hop is passed to the reasoning module that produces the final output. }
\label{fullmodel}
\end{center}
\vskip -0.2in
\end{figure*}

Many promising advances for achieving complex reasoning with neural networks have been obtained during the last years. 
Unlike symbolic approaches to complex reasoning, deep neural networks can learn representations from perceptual information. Because of that, they do not suffer from the symbol grounding problem ~\cite{symbol}, and can generalize better than classical symbolic approaches. Most of these neural network models make use of an explicit memory storage and an attention mechanism. For instance, Memory Networks (MemNN), Dynamic Memory Networks (DMN) or Neural Turing Machines (NTM)~\cite{memorynetworks,kumar2016ask, graves2014neural} build explicit memories from the perceptual inputs and access these memories using learned attention mechanisms. After that some memories have been attended, using a multi-step procedure, the attended memories are combined and passed through a simple output layer that produces a final answer. While this allows some multi-step inferential process, these networks lack a more complex reasoning mechanism, needed for more elaborated tasks such as inferring relations among entities (relational reasoning). On the contrary, Relation Networks (RNs), proposed in \citet{santoro2017simple}, have shown outstanding performance in relational reasoning tasks. Nonetheless, a major drawback of RNs is that they consider each of the input objects in pairs, having to process a quadratic number of relations. That limits the usability of the model on large problems and makes forward and backward computations quite expensive. To solve these problems we propose a novel Memory Network architecture called the Working Memory Network (W-MemNN). Our model augments the original MemNN with a relational reasoning module and a new working memory buffer. 

The attention mechanism of the Memory Network allows the filtering of irrelevant inputs, reducing a lot of the computational complexity while keeping the relational reasoning capabilities of the RN.
Three main components compose the W-MemNN: An input module that converts the perceptual inputs into an internal vector representation and save these representations into a short-term storage, an attentional controller that attend to these internal representations and update a working memory buffer, and a reasoning module that operates on the set of objects stored in the working memory buffer in order to produce a final answer. This component-based architecture is inspired by the well-known model from cognitive sciences called the multi-component working memory model, proposed in~\citet{baddeley1974working}.\\ 
We studied the proposed model on the text-based QA benchmark bAbI~\cite{weston2015towards} which consists of 20 different toy tasks that measure different reasoning skills. While models such as EntNet~\cite{henaff2016tracking} have focused on the per-task training version of the benchmark (where a different model is trained for each task), we decided to focus on the jointly trained version of the task, where the model is trained on all tasks simultaneously. In the jointly trained bAbI-10k benchmark we achieved state-of-the-art performance, improving the previous state-of-the-art on more than 2\%. Moreover, a simple ensemble of two of our models can solve all 20 tasks simultaneously. Also, we tested our model on the visual QA dataset NLVR. In that dataset, we obtained performance at the level of the Module Neural Networks~\cite{andreas2016neural}. Our model, however, achieves these results using the raw input statements, without the extra text processing used in the Module Networks.

Finally, qualitative and quantitative analysis shows that the inclusion of the Relational Reasoning module is crucial to improving the performance of the MemNN on tasks that involve relational reasoning. We can achieve this performance by also reducing the computation times of the RN considerably. Consequently, we hope that this contribution may allow applying RNs to larger problems.

\section{Model}
Our model is based on the Memory Network architecture. Unlike MemNN we have included a reasoning module that helps the network to solve more complex tasks. The proposed model consists of three main modules: An input module, an attentional controller, and a reasoning module. The model processes the input information in multiple passes or hops. At each pass the output of the previous hop can condition the current pass, allowing some incremental refinement.\\
\textbf{Input module: } The input module converts the perceptual information into an internal feature representation. The input information can be processed in chunks, and each chunk is saved into a short-term storage. The definition of what is a chunk of information depends on each task. For instance, for textual question answering, we define each chunk as a sentence. Other options might be n-grams or full documents. This short-term storage can only be accessed during the hop.\\
\textbf{Attentional Controller: }
The attentional controller decides in which parts of the short-term storage the model should focus. The attended memories are kept during all the hops in a working memory buffer. The attentional controller is conditioned by the task at hand, for instance, in question answering the question can condition the attention. Also, it may be conditioned by the output of previous hops, allowing the model to change its focus to new portions of the memory over time.\\
Many models compute the attention for each memory using a compatibility function between the memory and the question. Then, the output is calculated as the weighted sum of the memory values, using the attention as weight. A simple way to compute the attention for each memory is to use dot-product attention. This kind of mechanism is used in the original Memory Network and computes the attention value as the dot product between each memory and the question. Although this kind of attention is simple, it may not be enough for more complex tasks. Also, since there are no learned weights in the attention mechanism, the attention relies entirely on the learned embeddings. That is something that we want to avoid in order to separate the learning of the input and attention module. One way to allow learning in the dot-product attention is to project the memories and query vectors linearly. This is done by multiplying each vector by a learned projection matrix (or equivalently a feed-forward neural network). In this way, we can set apart the attention and input embeddings learning, and also allow more complex patterns of attention.

\textbf{Reasoning Module: } The memories stored in the working memory buffer are passed to the reasoning module. The choice of reasoning mechanism is left open and may depend on the task at hand. In this work, we use a Relation Network as the reasoning module. The RN takes the attended memories in pairs to infer relations among the memories. This can be useful, for example, in tasks that include comparisons.\\
A detailed description of the full model is shown in Figure~\ref{fullmodel}.
\subsection{W-MemN2N for Textual Question Answering}
We proceed to describe an implementation of the model for textual question answering.
In textual question answering the input consists of a set of sentences or facts, a question, and an answer. The goal is to answer the question correctly based on the given facts.\\
Let $(s, q, a)$ represents an input sample, consisting of a set of sentences $s = \{x_i\}_{i=1}^L$, a query $q$ and an answer $a$. Each sentence contains $M$ words,  $\{w_i\}_{i=1}^M$, where each word is represented as a one-hot vector of length $|V|$, being $|V|$ the vocabulary size. The question contains $Q$ words, represented as in the input sentences.
\subsubsection*{Input Module}
Each word in each sentence is encoded into a vector representation $v_i$ using an embedding matrix $W \in \mathbb{R}^{|V|\times d}$, where $d$ is the embedding size. Then, the sentence is converted into a memory vector $m_i$ using the final output of a gated recurrent neural network (GRU)~\cite{chung2014empirical}:
\begin{eqnarray*}
m_i = \gru([v_1, v_2, ..., v_M])
\end{eqnarray*}
Each memory $\{m_i\}_{i=1}^L$, where $m_i \in \mathbb{R}^d$, is stored into the short-term memory storage. The question is encoded into a vector $u$ in a similar way,  using the output of a gated recurrent network. 
\subsubsection*{Attentional Controller}
Our attention module is based on the Multi-Head attention mechanism proposed in \citet{vaswani}. First, the memories are projected using a projection matrix $W_m \in \mathbb{R}^{d \times d}$, as $m_i' = W_m m_i$.
Then, the similarity between the projected memory and the question is computed using the Scaled Dot-Product attention:
\begin{eqnarray}
\alpha_i &=& \softmax \big(\frac{u^Tm_i'}{\sqrt{d}}\big)\\
       &=& \frac{\exp((u^Tm_i')/\sqrt{d})}{\sum_j \exp((u^Tm_j')/\sqrt{d})}.
\label{scaled-attention}
\end{eqnarray}
Next, the memories are combined using the attention weights $\alpha_i$, obtaining an output vector $h = \sum_{j} \alpha_j m_j$.\\ 
In the Multi-Head mechanism, the memories are projected $S$ times using different projection matrices $\{W^s_m\}_{s=1}^S$. For each group of projected memories, an output vector $\{h_i\}_{i=1}^S$ is obtained using the Scaled Dot-Product attention~(eq.~\ref{scaled-attention}). Finally, all vector outputs are concatenated and projected again using a different matrix:
\begin{eqnarray*}
o_k = [h_1; h_2; ...;h_S] W_o,
\end{eqnarray*}
where ; is the concatenation operator and $W_o \in \mathbb{R}^{Sd \times d}$. The $o_k$ vector is the final response vector for the hop $k$. This vector is stored in the working memory buffer. The attention procedure can be repeated many times (or hops). At each hop, the attention can be conditioned on the previous hop by replacing the question vector $u$ by the output of the previous hop. To do that we pass the output through a simple neural network $f_t$. Then, we use the output of the network as the new conditioner:
\begin{eqnarray}
o^n_k = f_t(o_k).
\end{eqnarray}
This network allows some learning in the transition patterns between hops.\\
We found Multi-Head attention to be very useful in the joint bAbI task. This can be a product of the intrinsic multi-task nature of the bAbI dataset. A possibility is that each attention head is being adapted for different groups of related tasks. However, we did not investigate this further.\\
Also, note that while in this section we use the same set of memories at each hop, this is not necessary. For larger sequences each hop can operate in different parts of the input sequence, allowing the processing of the input in various steps.
\subsubsection*{Reasoning Module}
The outputs stored in the working memory buffer are passed to the reasoning module. The reasoning module used in this work is a Relation Network (RN). In the RN the output vectors are concatenated in pairs together with the question vector. Each pair is passed through a neural network $g_\theta$ and all the outputs of the network are added to produce a single vector. Then, the sum is passed to a final neural network $f_\phi$:
\begin{eqnarray}
r = f_\phi\bigg(\sum_{i,j} g_\theta ([o_i; o_j; u])\bigg),
\end{eqnarray}
The output of the Relation Network is then passed through a final weight matrix and a softmax to produce the predicted answer:
\begin{eqnarray}
\hat{a} = \softmax(Vr),
\end{eqnarray}
where $V \in \mathbb{R}^{|A|\times d_\phi}$, $|A|$ is the number of possible answers and $d_\phi$ is the dimension of the output of $f_\phi$.
The full network is trained end-to-end using standard cross-entropy between $\hat{a}$ and the true label $a$.
\section{Related Work}
\subsection{Memory Augmented Neural Networks}
During the last years, there has been plenty of work on achieving complex reasoning with deep neural networks. An important part of these developments has used some kind of explicit memory and attention mechanisms. One of the earliest recent work is that of Memory Networks~\cite{memorynetworks}. Memory Networks work by building an addressable memory from the inputs and then accessing those memories in a series of reading operations. Another, similar, line of work is the one of Neural Turing Machines. They were proposed in~\citet{graves2014neural} and are the basis for recent neural architectures including the Differentiable Neural Computer (DNC) and the Sparse Access Memory (SAM)~\cite{graves2016hybrid, rae2016scaling}. The NTM model also uses a content addressable memory, as in the Memory Network, but adds a write operation that allows updating the memory over time. The management of the memory, however, is different from the one of the MemNN. While the MemNN model pre-load the memories using all the inputs, the NTM writes and read the memory one input at a time. 

An additional model that makes use of explicit external memory is the Dynamic Memory Network (DMN)~\cite{kumar2016ask,xiong2016dynamic}. The model shares some similarities with the Memory Network model. However, unlike the MemNN model, it operates in the input sequentially (as in the NTM model). The model defines an Episodic Memory module that makes use of a Gated Recurrent Neural Network (GRU) to store and update an internal state that represents the episodic storage. 

\subsection{Memory Networks}
Since our model is based on the MemNN architecture, we proceed to describe it in more detail. The Memory Network model was introduced in~\citet{memorynetworks}. In that work, the authors proposed a model composed of four components: The input feature map that converts the input into an internal vector representation, the generalization module that updates the memories given the input, the output feature map that produces a new output using the stored memories, and the response module that produces the final answer. The model, as initially proposed, needed some strong supervision that explicitly tells the model which memories to attend. In order to solve that limitation, the End-To-End Memory Network (MemN2N) was proposed in~\citet{sukhbaatar2015end}.

The model replaced the hard-attention mechanism used in the original MemNN by a soft-attention mechanism that allowed to train it end-to-end without strong supervision. In our model, we use a component-based approach, as in the original MemNN architecture. However, there are some differences: First, our model makes use of two external storages: a short-term storage, and a working memory buffer. The first is equivalent to the one updated by the input and generalization module of the MemNN. The working memory buffer, on the other hand, does not have a counterpart in the original model. Second, our model replaces the response module by a reasoning module. Unlike the original MemNN, our reasoning module is intended to make more complex work than the response module, that was only designed to produce a final answer.

\subsection{Relation Networks}
The ability to infer and learn relations between entities is fundamental to solve many complex reasoning problems. Recently, a number of neural network models have been proposed for this task. These include Interaction Networks, Graph Neural Networks, and Relation Networks~\cite{battaglia2016interaction, scarselli2009graph, santoro2017simple}. In specific, Relation Networks (RNs) have shown excellent results in solving textual and visual question answering tasks requiring relational reasoning. The model is relatively simple: First, all the inputs are grouped in pairs and each pair is passed through a neural network. Then, the outputs of the first network are added, and another neural network processes the final vector. The role of the first network is to infer relations among each pair of objects. In~\citet{recurrentrelationalreasoning} the authors propose a recurrent extension to the RN. By allowing multiple steps of relational reasoning, the model can learn to solve more complex tasks. The main issue with the RN architecture is that its scale very poorly for larger problems. This is because it operates on $O(n^2)$ pairs, where $n$ is the number of input objects (for instance, sentences in the case of textual question answering). Something that becomes quickly prohibitive for tasks involving many input objects.

\subsection{Cognitive Science}
The concept of working memory has been extensively developed in cognitive psychology. It consists of a limited capacity system that allows temporary storage and manipulation of information and is crucial to any reasoning task. One of the most influential models of working memory is the multi-component model of working memory proposed by~\citet{baddeley1974working}. This model is composed both of a supervisory attentional controller (the central executive) and two short-term storage systems: The phonological loop, capable of holding speech-based information, and the visuospatial sketchpad, concerned with visual storage. The central executive plays various functions, including the capacity to focus attention, to divide attention and to control access to long-term memory. Later modifications to the model~\cite{baddeley2000episodic} include an episodic buffer that is capable of integrating and holding information from different sources. Connections of the working memory model to memory augmented neural networks have been already studied in~\citet{graves2014neural}. We follow this effort and subdivide our model into components that resemble (in a basic way) the multi-component model of working memory. Note, however, that we use the term working memory buffer instead of episodic buffer. This is because the episodic buffer has an integration function that our model does not cover. However, that can be an interesting source of inspiration for next versions of the model that integrate both visual and textual information for question answering.

\begin{table*}[t]
    \centering
      \setlength{\tabcolsep}{10pt}
    \begin{tabular}{l|c|c|c|c|c|c}

 & LSTM & MN-S & MN & SDNC & WMN & WMN$^\dag$ \\
\hline
 1:    1 supporting fact  & $ 0.0 $ & $  0.0$ & $  0.0$ & $ 0.0 $ &  $ 0.0 $ & $ 0.0 $\\
 2:   2 supporting facts  & $ 81.9$ & $  0.0$ & $  1.0$ & $ 0.6 $ &  $ 0.7 $ & $ 0.3 $\\
 3:   3 supporting facts  & $ 83.1$ & $  0.0$ & $  6.8$ & $ 0.7 $ &  $ 5.3 $ & $ 4.6 $\\
 4: 2 argument relations  & $ 0.2 $ & $  0.0$ & $  0.0$ & $ 0.0 $ &  $ 0.0 $ & $ 0.0 $\\
 5: 3 argument relations  & $ 1.2 $ & $  0.3$ & $  6.1$ & $ 0.3 $ &  $ 0.6 $ & $ 0.4 $\\
 6:     yes/no questions  & $ 51.8$ & $  0.0$ & $  0.1$ & $ 0.0 $ &  $ 0.0 $ & $ 0.0 $\\
 7:             counting  & $ 24.9$ & $  3.3$ & $  6.6$ & $ 0.2 $ &  $ 0.6 $ & $ 0.5 $\\
 8:           lists/sets  & $ 34.1$ & $  1.0$ & $  2.7$ & $ 0.2 $ &  $ 0.2 $ & $ 0.3 $\\
 9:      simple negation  & $ 20.2$ & $  0.0$ & $  0.0$ & $ 0.0 $ &  $ 0.0 $ & $ 0.0 $\\
10: indefinite knowledge  & $ 30.1$ & $  0.0$ & $  0.5$ & $ 0.2 $ &  $ 0.5 $ & $ 0.0 $\\
11:    basic coreference  & $ 10.3$ & $  0.0$ & $  0.0$ & $ 0.0 $ &  $ 0.3 $ & $ 0.0 $\\
12:          conjunction  & $ 23.4$ & $  0.0$ & $  0.1$ & $ 0.1 $ &  $ 0.0 $ & $ 0.0 $\\
13: compound coreference  & $ 6.1 $ & $  0.0$ & $  0.0$ & $ 0.1 $ &  $ 0.0 $ & $ 0.0 $\\
14:       time reasoning  & $ 81.0$ & $  0.0$ & $  0.0$ & $ 0.1 $ &  $ 0.0 $ & $ 0.0 $\\
15:      basic deduction  & $ 78.7$ & $  0.0$ & $  0.2$ & $ 0.0 $ &  $ 0.0 $ & $ 0.0 $\\
16:      basic induction  & $ 51.9$ & $  0.0$ & $  0.2$ & $ 54.1$ &  $ 0.0 $ & $ 0.3 $\\
17: positional reasoning  & $ 50.1$ & $  24.6$ & $ 41.8$ & $ 0.3 $ & $ 0.3 $ & $ 0.1 $\\
18:       size reasoning  & $ 6.8 $ & $ 2.1$ & $  8.0$ & $ 0.1 $   & $ 0.1 $ & $ 0.4 $\\
19:         path finding  & $ 90.3$ & $  31.9$ & $ 75.7$ & $ 1.2 $ & $ 0.6 $ & $ 0.0 $\\
20:  agent's motivations  & $ 2.1 $ & $ 0. $ & $  0.0$ & $ 0.0 $ & $ 0.0 $ & $ 0.0 $\\
\hline
Mean Error (\%)           & $36.4 $ & $ 3.2$ &  $  7.5$ &  $2.8 $ &  $ 0.4$ &  $ \textbf{0.3}$\\
Failed tasks (err. $>$ 5\%) & $16 $ & $    2$ &   $    6$ &  $1   $ &  $  1  $ & $ \textbf{0}  $\\
    \end{tabular}
    \caption{Test accuracies on the jointly trained bAbI-10k dataset. \texttt{MN-S} stands for strongly supervised Memory Network, \texttt{MN-U} for end-to-end Memory Network without supervision, and \texttt{WMN} for Working Memory Network. Results for \texttt{{LSTM}}, \texttt{MN-U}, and \texttt{MN-S} are taken from~\citet{sukhbaatar2015end}. Results for \texttt{SDNC} are taken from~\citet{rae2016scaling}. \texttt{WMN$^\dag$} is an ensemble of two Working Memory Networks.}
    \label{babiresults}
\end{table*}

\section{Experiments}

\subsection{Textual Question Answering}
To evaluate our model on textual question answering we used the Facebook bAbI-10k dataset~\cite{weston2015towards}.
The bAbI dataset is a textual QA benchmark composed of 20 different tasks. Each task is designed to test a different reasoning skill, such as deduction, induction, and coreference resolution. Some of the tasks need relational reasoning, for instance, to compare the size of different entities. Each sample is composed of a question, an answer, and a set of facts. There are two versions of the dataset, referring to different dataset sizes: bAbI-1k and bAbI-10k. In this work, we focus on the bAbI-10k version of the dataset which consists of $10,000$ training samples per task. A task is considered solved if a model achieves greater than $95\%$ accuracy. Note that training can be done per-task or joint (by training the model on all tasks at the same time). Some models~\cite{liu2017gated} have focused in the per-task training performance, including the EntNet model~\cite{henaff2016tracking} that solves all the tasks in the per-task training version. We choose to focus on the joint training version since we think is more indicative of the generalization properties of the model. A detailed analysis of the dataset can be found in~\citet{lee2015reasoning}.

\subsubsection*{Model Details}

To encode the input facts we used a word embedding that projected each word in a sentence into a real vector of size $d$. We defined $d=30$ and used a GRU with 30 units to process each sentence. We used the 30 sentences in the support set that were immediately prior to the question. The question was processed using the same configuration but with a different GRU. We used 8 heads in the Multi-Head attention mechanism.
For the transition network $f_t$, which operates in the output of each hop, we used a two-layer MLP consisting of 15 and 30 hidden units (so the output preserves the memory dimension). We used $H=4$ hops (or equivalently, a working memory buffer of size 4). In the reasoning module, we used a 3-layer MLP consisting of 128 units in each layer and with ReLU non-linearities for $g_\theta$. We omitted the $f_\phi$ network since we did not observe improvements when using it. The final layer was a linear layer that produced logits for a softmax over the answer vocabulary. 

\subsubsection*{Training Details}
We trained our model end-to-end with a cross-entropy loss function and using the Adam optimizer~\cite{kingma2014adam}. We used a learning rate of $\nu = 1e^{-3}$. We trained the model during 400 epochs. For training, we used a batch size of 32. As in~\citet{sukhbaatar2015end} we did not average the loss over a batch. Also, we clipped gradients with norm larger than 40~\cite{pascanu2013difficulty}. For all the dense layers we used $\ell_2$ regularization with value $1e^{-3}$. All weights were initialized using Glorot normal initialization~\cite{glorot2010understanding}. $10\%$ of the training set was held-out to form a validation set that we used to select the architecture and for hyperparameter tunning. In some cases, we found useful to restart training after the 400 epochs with a smaller learning rate of $1e^{-5}$ and anneals every 5 epochs by $\nu/2$ until 20 epochs were reached.

\subsubsection*{bAbI-10k Results}
On the jointly trained bAbI-10k dataset our best model (out of 10 runs) achieves an accuracy of 99.58\%. That is a 2.38\% improvement over the previous state-of-the-art that was obtained by the Sparse Differential Neural Computer (SDNC)~\cite{rae2016scaling}. The best model of the 10 runs solves almost all tasks of the bAbI-10k dataset  (by a 0.3\% margin). However, a simple ensemble of the best two models solves all 20 tasks and achieves an almost perfect accuracy of 99.7\%. We list the results for each task in Table~\ref{babiresults}. 
Other authors have reported high variance in the results, for instance, the authors of the SDNC report a mean accuracy and standard deviation over 15 runs of $93.6\pm 2.5$ (with $15.9 \pm 1.6$ passed tasks). In contrast, our model achieves a mean accuracy of $98.3\pm1.2$ (with $18.6 \pm 0.4$ passed tasks), which is better and more stable than the average results obtained by the SDNC.\\
The Relation Network solves 18/20 tasks. We achieve even better performance, and with considerably fewer computations, as is explained in Section~\ref{sec:on2}. We think that by including the attention mechanism, the relation reasoning module can focus on learning the relation among relevant objects, instead of learning spurious relations among irrelevant objects. For that, the Multi-Head attention mechanism was very helpful.\\

\subsubsection*{The Effect of the Relational Reasoning Module}

When compared to the original Memory Network, our model substantially improves the accuracy of tasks 17 (positional reasoning) and 19 (path finding). Both tasks require the analysis of multiple relations~\cite{lee2015reasoning}. For instance, the task 19 needs that the model reasons about the relation of different positions of the entities, and in that way find a path to arrive from one to another. The accuracy improves in 75.1\% for task 19 and in 41.5\% for task 17 when compared with the MemN2N model. Since both tasks require reasoning about relations, we hypothesize that the relational reasoning module of the W-MemNN was of great help to improve the performance on both tasks.\\
The Relation Network, on the other hand, fails in the tasks 2 (2 supporting facts) and 3 (3 supporting facts). Both tasks require handling a significant number of facts, especially in task 3. In those cases, the attention mechanism is crucial to filter out irrelevant facts.

\begin{table*}[tb]
\resizebox{0.5\textwidth}{!}{
\includegraphics[width=0.9\textwidth, trim={0 1cm 0 0}, clip]{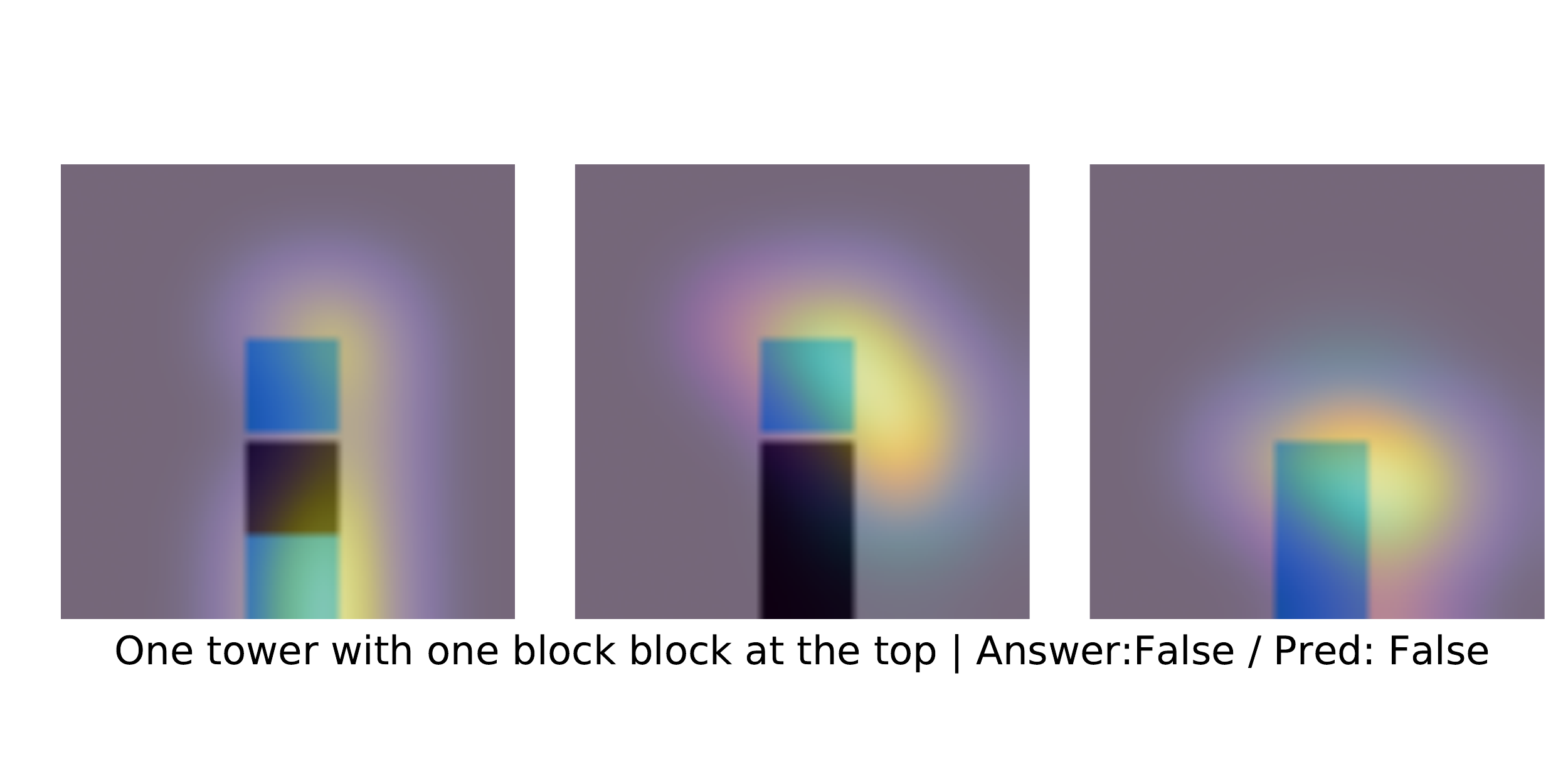}
}
\resizebox{0.5\textwidth}{!}{
\includegraphics[width=0.9\textwidth, trim={0 1cm 0 0}, clip]{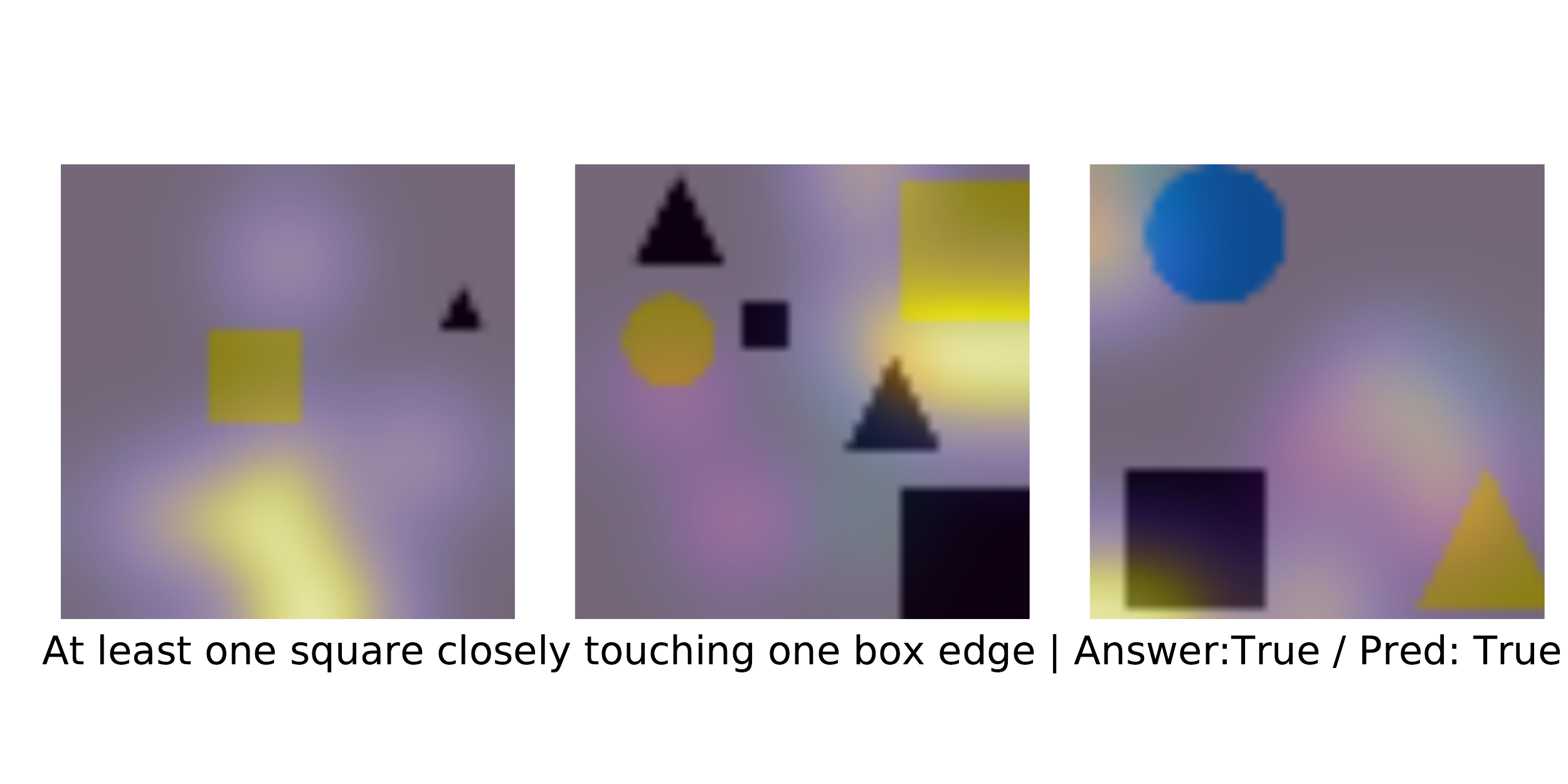}
}\\
\resizebox{0.525\textwidth}{!}{
\begin{tabular}{|lc|cccc|}
\hline
Story (2 supporting facts)& Support & Hop 1 & Hop 2 & Hop 3 & Hop 4\\
\hline
Mary moved to the office. & & \cellcolor{blue!19}0.79 & \cellcolor{blue!7}0.30 & \cellcolor{blue!3}0.15 & \cellcolor{blue!3}0.15\\
Sandra travelled to the bedroom. & True & \cellcolor{blue!0}0.02 & \cellcolor{blue!66}2.64 & \cellcolor{blue!68}2.75 & \cellcolor{blue!9}0.39\\
Daniel dropped the football. & & \cellcolor{blue!0}0.03 & \cellcolor{blue!3}0.13 & \cellcolor{blue!3}0.16 & \cellcolor{blue!10}0.41\\
Sandra left the milk there. & True & \cellcolor{blue!25}1.01 & \cellcolor{blue!1}0.07 & \cellcolor{blue!4}0.16 & \cellcolor{blue!9}0.38\\
Daniel grabbed the football there. & & \cellcolor{blue!2}0.08 & \cellcolor{blue!7}0.31 & \cellcolor{blue!1}0.07 & \cellcolor{blue!6}0.27\\
\hline
\multicolumn{6}{|l|}{\specialcell{Question: Where is the milk? Answer: bedroom, Pred: {\color{blue} bedroom}}}\\
\hline
\end{tabular}
}
\resizebox{0.475\textwidth}{!}{
\begin{tabular}{|lc|cccc|}
\hline
Story (basic induction)& Support & Hop 1 & Hop 2 & Hop 3 & Hop 4\\
\hline
Brian is white. & & \cellcolor{blue!11}0.46 & \cellcolor{blue!8}0.36 & \cellcolor{blue!8}0.35 & \cellcolor{blue!22}0.89\\
Bernhard is white. & & \cellcolor{blue!1}0.07 & \cellcolor{blue!3}0.13 & \cellcolor{blue!4}0.19 & \cellcolor{blue!20}0.81\\
Julius is a frog. & True & \cellcolor{blue!3}0.16 & \cellcolor{blue!50}2.03 & \cellcolor{blue!9}0.39 & \cellcolor{blue!6}0.26\\
Julius is white. & True & \cellcolor{blue!2}0.09 & \cellcolor{blue!5}0.23 & \cellcolor{blue!60}2.42 & \cellcolor{blue!32}1.32\\
Greg is a frog. & True & \cellcolor{blue!48}1.95 & \cellcolor{blue!40}1.60 & \cellcolor{blue!19}0.77 & \cellcolor{blue!6}0.25\\
\hline
\multicolumn{6}{|l|}{\specialcell{Question: What color is greg? Answer: white, Pred: {\color{blue} white}}}\\
\hline
\end{tabular}
}
\caption{Examples of visualizations of attention for textual and visual QA. Top: Visualization of attention values for the NLVR dataset. To obtain more aesthetic figures we applied a gaussian blur to the attention matrix. Bottom: Attention values for the bAbI dataset. In each cell, the sum of the attention for all heads is shown. \label{tbl:memnn2}
}
\end{table*}

\subsection{Visual Question Answering}
To further study our model we evaluated its performance on a visual question answering dataset. For that, we used the recently proposed NLVR dataset~\cite{P17-2034}. Each sample in the NLVR dataset is composed of an image with three sub-images and a statement. The task consists in judging if the statement is true or false for that image. Evaluating the statement requires reasoning about the sets of objects in the image, comparing objects properties, and reasoning about spatial relations. The dataset is interesting for us for two reasons. First, the statements evaluation requires complex relational reasoning about the objects in the image. Second,  unlike the bAbI dataset, the statements are written in natural language. Because of that, each statement displays a range of syntactic and semantic phenomena that are not present in the bAbI dataset. 

\subsubsection*{Model details}

Our model can be easily adapted to deal with visual information. Following the idea from~\citet{santoro2017simple}, instead of processing each input using a recurrent neural network, we use a Convolutional Neural Network (CNN). The CNN takes as input each sub-image and convolved them through convolutional layers. The output of the CNN consists of $k$ feature maps (where $k$ is the number of kernels in the final convolutional layer) of size $d \times d$. Then, each memory is built from the vector composed by the concatenation of the cells in the same position of each feature map. Consequently, $d \times d$ memories of size $k$ are stored in the short-term storage. The statement is processed using a GRU neural network as in the textual reasoning task. Then, we can proceed using the same architecture for the reasoning and attention module that the one used in the textual QA model. However, for the visual QA task, we used an additive attention mechanism. The additive attention computes the attention weight using a feed-forward neural network applied to the concatenation of the memory vector and statement vector.  

\subsubsection*{Results}
Our model achieves a validation / test accuracy of $65.6\%/65.8\%$. Notably, we achieved a performance comparable to the results of the Module Neural Networks~\cite{andreas2016neural} that make use of standard NLP tools to process the statements into structured representations. Unlike the Module Neural Networks, we achieved our results using only raw input statements, allowing the model to learn how to process the textual input by itself. Given the more complex nature of the language used in the NLVR dataset we needed to use a larger embedding size and GRU hidden layer than in the bAbI dataset (100 and 128 respectively). That, however, is a nice feature of separating the input from the reasoning and attention component: One way to process more complex language statements is increasing the capacity of the input module.

\subsection{From $O(n^2)$ to $O(n)$}
\label{sec:on2}
\begin{figure}[ht]
\vskip 0.2in
\begin{center}
\centerline{\includegraphics[width=0.9\columnwidth]{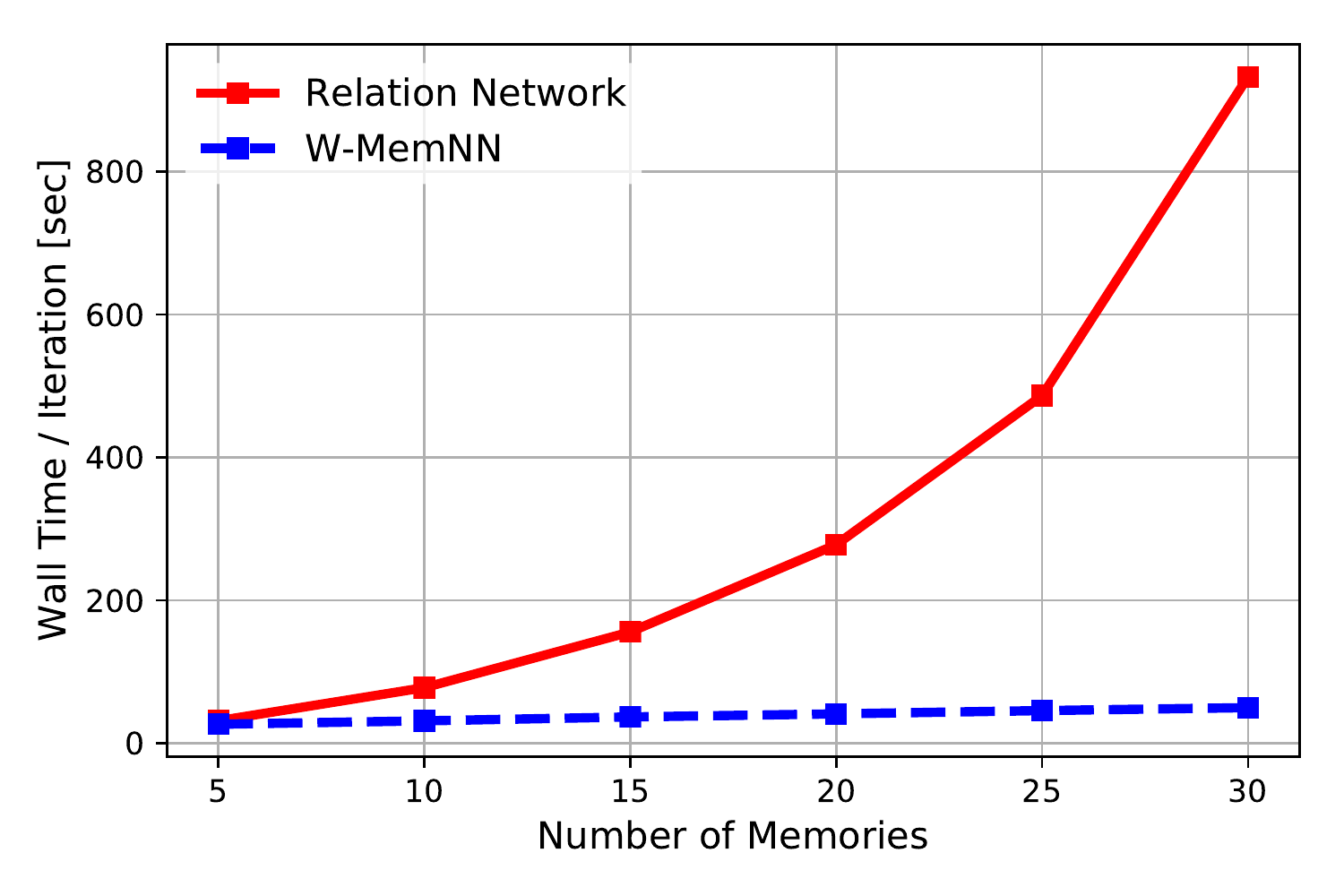}}
\caption{Wall-clock times for a forward and backward pass for a single batch. The batch size used is 32. While for 5 memories the times are comparable, for 30 memories the W-MemNN takes around 50s while the RN takes 930s, a speedup of almost $20\times$.}
\label{time_comparison}
\end{center}
\vskip -0.2in
\end{figure}

One of the major limitations of RNs is that they need to process each one of the memories in pairs. To do that, the RN must perform $O(n^2)$ forward and backward passes (where $n$ is the number of memories). This becomes quickly prohibitive for a larger number of memories. In contrast, the dependence of the W-MemNN run times on the number of memories is linear. Note, however, that computation times in the W-MemNN depend quadratically on the size of the working memory buffer. Nonetheless, this number is expected to be much smaller than the number of memories. To compare both models we measured the wall-clock time for a forward and backward pass for a single batch of size 32. We performed these experiments on a GPU NVIDIA K80. Figure~\ref{time_comparison} shows the results. 

\subsection{Memory Visualizations}

One nice feature from Memory Networks is that they allow some interpretability of the reasoning procedure by looking at the attention weights. At each hop, the attention weights show which parts of the memory the model found relevant to produce the output. RNs, on the contrary, lack of this feature. Table~\ref{tbl:memnn2} shows the attention values for visual and textual question answering.

\section{Conclusion}

We have proposed a novel Working Memory Network architecture that introduces improved reasoning abilities to the original MemNN model. We demonstrated that by augmenting the MemNN architecture with a Relation Network, the computational complexity of the RN can be reduced, without loss of performance. This opens the opportunity for using RNs in larger problems, something that may be very useful, given the many tasks requiring a significant amount of memories.\\
Although we have used RN as the reasoning module in this work, other options can be tested. It might be interesting to analyze how other reasoning modules can improve different weaknesses of the model.\\
We presented results on the jointly trained bAbI-10k dataset, where we achieve a new state-of-the-art, with an average error of less than 0.5\%. Also, we showed that our model can be easily adapted for visual question answering.\\
Our architecture combines perceptual input processing, short-term memory storage, an attention mechanism, and a reasoning module. While other models have focused on different parts of these components, we think that is important to find ways to combine these different mechanisms if we want to build models capable of complex reasoning. Evidence from cognitive sciences seems to show that all these abilities are needed in order to achieve human-level complex reasoning.

\section*{Acknowledgments}
JP was supported by the Scientific and Technological Center of Valpara\'iso (CCTVal) under Fondecyt grant BASAL FB0821. HA was supported through the research project Fondecyt-Conicyt 1170123.  The work of HAC was supported by the research project Fondecyt Initiation into Research 11150248.
\bibliography{acl2018}
\bibliographystyle{acl_natbib}

\appendix

\end{document}